\documentclass[conference]{IEEEtran}
\IEEEoverridecommandlockouts

\usepackage{cite}
\usepackage{amsthm,amsmath,amssymb,amsfonts}

\usepackage{algorithmic}
\usepackage{algorithm}
\usepackage{graphicx}
\usepackage{textcomp}
\usepackage{xcolor}
\def\BibTeX{{\rm B\kern-.05em{\sc i\kern-.025em b}\kern-.08em
    T\kern-.1667em\lower.7ex\hbox{E}\kern-.125emX}}

\usepackage{subfloat}
\usepackage{subfig}

\makeatletter
\newcommand{\linebreakand}{%
    \end{@IEEEauthorhalign}
    \hfill\mbox{}\par
    \mbox{}\hfill\begin{@IEEEauthorhalign}
}
\makeatother

\newtheorem{definition}{Definition}

\begin{document}

\title{Analyzing Neural Network Robustness Using Graph Curvature\thanks{This paper is a pre-print version of the published paper~\cite{tan24} (note that some numbers in Table~\ref{tab:all_auc} were updated after a small implementation fix). This material is based upon work supported by the National Science Foundation (NSF) under Award No. 2403616, as part of the NSF Cyber-Physical Systems Program. Any opinions, findings and conclusions or recommendations expressed in this material are those of the author(s) and do not necessarily reflect the views of the National Science Foundation.}
}

\author{\IEEEauthorblockN{Shuhang Tan}
\IEEEauthorblockA{\textit{Computer Science Department} \\
\textit{Rensselaer Polytechnic Institute}\\
Troy, NY, USA \\
tans5@rpi.edu}
\and
\IEEEauthorblockN{Jayson Sia~~~~~~~~~~~~~~Paul Bogdan}
\IEEEauthorblockA{\textit{Dept. of Electrical and Computer Engineering} \\
\textit{University of Southern California}\\
Los Angeles, CA, USA \\
\{jsia, pbogdan\}@usc.edu}
\and
\IEEEauthorblockN{Radoslav Ivanov}
\IEEEauthorblockA{\textit{Computer Science Department} \\
\textit{Rensselaer Polytechnic Institute}\\
Troy, NY, USA \\
ivanor@rpi.edu}
}

\maketitle

\begin{abstract}
This paper presents a new look at the neural network (NN) robustness problem, from the point of view of graph theory analysis, specifically graph curvature. Graph curvature (e.g., Ricci curvature) has been used to analyze system dynamics and identify bottlenecks in many domains, including road traffic analysis and internet routing. We define the notion of neural Ricci curvature and use it to identify bottleneck NN edges that are heavily used to ``transport data" to the NN outputs. We provide an evaluation on MNIST that illustrates that such edges indeed occur more frequently for inputs where NNs are less robust. These results will serve as the basis for an alternative method of robust training, by minimizing the number of bottleneck edges.
\end{abstract}


\section{Introduction}
\label{sec:intro}
Autonomous systems (AS) increasingly use neural networks (NNs) due to their ability to process high-dimensional data such as camera images~\cite{dosovitskiy20}, LiDAR scans~\cite{ivanov2020b} and textual prompts~\cite{driess23}. At the same time, NNs are known to suffer from robustness vulnerabilities: a slightly perturbed or out-of-distribution input~\cite{szegedy13,recht19} may lead to very different and unexpected outputs. In turn, such vulnerabilities may severely compromise the safety and predictability of NN-based AS.

Since the discovery of NN robustness issues~\cite{szegedy13}, there has been an impressive amount of research on this topic. Researchers have developed a number of robust training methods, including adversarial training~\cite{madry17}, certified robustness~\cite{cohen19,wong18}, knowledge distillation~\cite{papernot16}, and semi-infinite constrained learning~\cite{robey21}. Although significant progress has been made, training robust NNs remains largely an unsolved and very challenging problem (e.g., the current leader on the CIFAR-10 robustness leaderboard~\cite{robust-bench} can only achieve high robust accuracy for perturbations of at most 8/255).

We note that the vast majority of existing methods approach the problem from an optimization point of view: e.g., in adversarial training the goal is to train a NN that minimizes the loss not only on training data but also on the worst-case bounded perturbations of that data. This bilevel non-convex optimization problem is challenging to solve and leads to suboptimal solutions, especially if gradient descent is used.

We take a fresh look at NN robustness through the lens of graph theory and network science analysis, in particular graph curvature (GC). GC (e.g., Ricci curvature~\cite{ollivier2009ricci}) has been effectively applied in numerous domains that can be modeled as graphs, including road traffic analysis~\cite{wang22,gao19}, internet routing~\cite{ni2015ricci}, machine learning~\cite{znaidi2023unified,li2022curvature}, and biological networks~\cite{farooq2019network,znaidi2023unified}, due to its ability to capture intrinsic geometric and local structure of the space, such as connectivity and robustness in networks. GC can quantify the importance of specific edges; for example, an edge with \emph{negative curvature} can be considered a bottleneck and is greatly important for the overall graph functionality, e.g., such an edge may connect different communities within the graph~\cite{Sia2019, sia2022inferring}.


In this paper, we employ GC in order to analyze the robustness of NN classifiers. We introduce the notion of neural Ricci curvature (NRC) that captures the bottleneck intuition of standard Ricci curvature -- if an edge has a negative NRC, then it is heavily used by the NN and is thus likely a source of robustness vulnerability. To calculate the NRC, we construct a neural data graph, i.e., a graph in the shape of the NN architecture, where edges are weighted by a combination of the NN weights and the magnitude of data that goes through each edge when an example is provided as input.

We evaluate the significance of the NRC using NNs trained on MNIST. We show that neural data graphs corresponding to more robust examples (i.e., examples which are correctly classified even for an adversarial perturbation) indeed have fewer negative-NRC edges. The results are consistent across architectures, including adversarially trained ones. This result will serve as the basis for an alternative, graph-based, method for robust training, that minimizes the number of negative-NRC edges and promotes balanced usage of all NN edges.

In summary, this paper makes two contributions: 1) we define the concepts of \emph{neural data graphs} and \emph{neural Ricci curvature} that can be used to identify bottleneck NN edges that contribute to robustness issues; 2) we provide an evaluation on MNIST that demonstrates that bottleneck edges indeed occur more frequently in examples where NNs are less robust.
\section{Background}
\label{sec:background}
The concept of Ricci curvature~\cite{bochner46} is used in Riemannian geometry to quantify the degree to which the geometry of a space deviates from being flat, as is the case for Euclidean space. In continuous manifolds, positive curvature is seen in spherical surfaces where geodesics converge, negative curvature is found in hyperbolic surfaces where geodesics diverge, and zero curvature characterizes flat Euclidean surfaces with parallel geodesics. The Ollivier-Ricci curvature (ORC)~\cite{ollivier2009ricci} serves as a discrete analogue, e.g., in the case of graphs, to curvature measurements in continuous spaces, and it is computed using transport theory via the Wasserstein distance. 

\begin{definition}[Ollivier-Ricci Curvature~\cite{ollivier2009ricci}]
Given a graph $G(V, E)$ with vertex set $V$ and edge set $E$, the ORC $\kappa(u, v)$ between two adjacent vertices $u$ and $v$ is given by
\vspace{-3px}
\begin{equation}
    \kappa(u,v) = 1 - \frac{W_1(m_u, m_v)}{d(u,v)},
\end{equation}
\vspace{-2px}
where $d(u,v)$ is the shortest-path distance between $u$ and $v$, $m_u$ and $m_v$ are the probability distributions over the neighbors of $u$ and $v$, respectively; $W_1(m_u, m_v)$ is the Wasserstein distance between distributions $m_u$ and $m_v$ and is given by
\vspace{-3px}
\begin{align*}
    W_1(m_u, m_v) = \inf_{\mu_{u,v} \in \Pi_{u,v}} \sum_{(u', v') \in V \times V} d(u',v')\mu_{u,v}(u',v'),
\end{align*}
\vspace{-8px}
\\
where $\Pi_{u,v}$ is the set of probability measures $\mu_{u,v}$ that capture all possible ways of transferring mass from $m_u$ to $m_v$, i.e.,
\vspace{-3px}
\begin{align*}
    \sum_{v'\in V} \mu_{u,v}(u',v') = m_u(u'), \quad \sum_{u'\in V} \mu_{u,v}(u',v') = m_v (v').
\end{align*}
\vspace{-12px}
\end{definition}
For unweighted nodes, the probability distribution is typically distributed uniformly to all the neighbors of $u$ and $v$. Otherwise, the probability distribution is adjusted based on the edge weights, e.g., using an exponential function~\cite{ni19}.

A positive curvature indicates that nodes within a particular region are tightly connected (due to a smaller $W_1$), suggesting a robust community structure. A negative curvature points to areas with more dispersed connections, often signifying edges that act as bridges between different communities. For example, if an edge $e$ between $u$ and $v$ is a bottleneck, then all paths from $u$'s neighbors to $v$'s neighbors have to go through $e$; in this case, $\kappa(u,v) < 0$ since $W_1 > d(u,v)$.
This makes ORC particularly effective at identifying bottleneck edges that may introduce robustness issues, i.e., the graph functionality is not robust to removing such negative-curvature edges.
\section{Problem Statement}
\label{sec:problem}
Consider a trained NN classifier $f_\theta: \mathcal{X} \to \{1,\dots, N\}$, where $\mathcal{X} \subseteq \mathbb{R}^n$ is the set of all input examples, and each example is assigned a label from 1 to $N$. We assume $f_{\theta}$ is an $L$-layer feedforward NN (possibly including convolutional layer), parameterized by ${\theta = \{(W_1, b_1), \dots, (W_L, b_L)\}}$, where $W_i$ and $b_i$ are the weights and biases of layer $i$. Thus, each layer $i$ can be written as a function $f_i(x) = \sigma(W_ix + b_i)$, where $\sigma$ is the ReLU activation: $\sigma(x) = \max\{0, x\}$.

An example $(x,y)$ is $\varepsilon$-\emph{robust} if there exists no $\varepsilon$-bounded perturbation of $x$ that changes the label predicted by $f$, i.e.,
\vspace{-3px}
\begin{equation}
f_\theta(x) = f_\theta(x + \delta) = y, \forall \delta \in \mathbb{R}^n \; \|\delta\|_\infty \le \varepsilon,
\end{equation}
where $y$ is the true (unknown) label of $x$. An example is called $\varepsilon$-\emph{nonrobust} if there exists a $\delta$ that satisfies $\|\delta\|_\infty \le \varepsilon$ for which $f_\theta(x) \neq f_\theta(x + \delta)$. Robust test accuracy is the fraction of $\varepsilon$-robust examples in the test set. Although NN robustness is distinct from graph robustness, we hypothesize that non-robust NNs correspond to non-robust graphs.

The problem considered in this paper is to define the notions of a neural data graph $G_{f_\theta, x}$ and a corresponding neural Ricci curvature. In particular, the NRC definition must capture the notion of a bottleneck edge such that a NN with more negative-NRC edges is less robust to input perturbations.

\textbf{Notation.} We use $n_{l,i}$ to denote neuron $i$ in layer $l$ in the NN; $n_{l,i}(x)$ denotes the neuron's activation given NN input $x$. Similarly, $w_{l,jk}$ denotes the NN weight on the edge between $n_{l,j}$ and $n_{l+1,k}$, i.e., matrix entry $[W_l]_{kj}$. Given graph nodes $u$ and $v$, we use $w(u,v)$ to denote the weight of edge $(u,v)$.
\section{Neural Ricci Curvature}
\label{sec:approach}
Our approach is motivated by prior work on analyzing the robustness of road and transit traffic networks to individual segments~\cite{wang22,gao19,znaidi2023unified}. To model the problem as a graph, one can map each intersection to a node and each segment to an edge; the edge weight is determined by historical data, e.g., travel time per passenger. Thus, if an edge has a negative ORC, i.e., its cost is lower than the costs of paths through the node's neighbors, then this edge is likely a bottleneck.

The above example has obvious analogues and differences with respect to NNs. The main similarities are: 1)~a NN has a natural graph structure; 2)~NN edges can be considered as roads that carry data; 3)~NN weights can be thought of as edge weights such that a larger NN weight means more data gets through (i.e.,~less travel time). However, two NN characteristics distinguish it from the traffic network case: i)~the NN has non-linear activations which break the traffic analogy (different levels of traffic ``get through" depending on the input); ii)~the NN has positive and negative weights, which makes it impossible to apply the vanilla ORC analysis.

In what follows, we define the concepts of a neural data graph and neural Ricci curvature, through iteratively addressing the challenges above: 1) we start with a neural graph; 2) we address the non-linearity challenge through calculating each ReLU's phase (0 or linear) for the current input $x$; 3) we address the mixed-sign challenge through normalizing the NN weights for the current input $x$.
\vspace{-3px}
\begin{definition}[Neural Graph]
\label{def:neural_graph}
Given a NN $f_\theta$, we define a neural graph $G_{f_\theta}$ as follows:
\begin{itemize}
    \item each NN neuron becomes a node in $G_{f_\theta}$ and each NN edge becomes an edge in $G_{f_\theta}$;

    \item each edge $(n_{l,i},n_{l+1,j})$ is assigned weight $1/|w_{l,ij}|$.
\end{itemize}
\end{definition}

Definition~\ref{def:neural_graph} is based purely on the NN and thus cannot address the challenges mentioned above. The following construction provides an approach that alleviates those challenges.

\begin{definition}[Neural Data Graph]
\label{def:neural_data_graph}
Given a NN $f_\theta$ and an example $x$, we define a neural data graph $G_{f_\theta,x}$ as follows:
\begin{itemize}
    \item construct neural graph $G_{f_\theta}$ according to Definition~\ref{def:neural_graph};

    \item an edge $(n_{l,i},n_{l+1,j})$ is removed from $G_{f_\theta}$ if ${n_{l,i}(x) = 0}$, i.e., the ReLU is in its zero phase;

    \item if weights $w_{l,1j}, \dots, w_{l,Kj}$ (where $K$ is the number of neurons in layer $l$) going into node ${n_{l+1,j}(x) > 0}$ have mixed signs, normalize weights using Algorithm~\ref{alg:normalization}.
\end{itemize}
\end{definition}

\begin{figure}[t]
\vspace{-10px}
\begin{algorithm}[H]
\caption{Mixed-Sign Weights Normalization}
\label{alg:normalization}
\begin{algorithmic}[1]
\renewcommand{\algorithmicrequire}{\textbf{Input:}}
 \renewcommand{\algorithmicensure}{\textbf{Output:}}
 \REQUIRE Mixed-sign NN weights $w_{l,1j}, \dots, w_{l,Kj}$, neuron activations $n_{l,1}(x), \dots, n_{l,K}(x)$ \\//Assuming layer $l$ has $K$ neurons
 \ENSURE  Graph weights $w(n_{l,1}, n_{l+1,j})$, $\dots, w(n_{l,K}, n_{l+1,j})$
 \\
  \STATE $sum = \sum_{i=1}^K w_{l,ij}n_{l,i}(x)$
  \STATE //it must be the case that $sum \ge 0$; otherwise ReLU would be in 0 phase
  \FOR {$i = 1$ to $K$}
  \IF {$w_{l,ij} < 0$}
  \STATE $w(n_{l,i}, n_{l+1,j}) = 0$
  \ELSE 
  \STATE $pos\_sum =\sum_{i=1}^K 1_{w_{l,1j} > 0}w_{l,1j}n_{l,1}(x)$
  \STATE $\hat{w}_{l,ij} = w_{l,ij}\frac{sum}{pos\_sum}$
  \STATE $w(n_{l,i}, n_{l+1,j}) = \frac{1}{\hat{w}_{l,ij}}$
  \ENDIF
  \ENDFOR
\end{algorithmic}
\end{algorithm}
\vspace{-20px}
\end{figure}

The last two parts of Definition~\ref{def:neural_data_graph} modify the original neural graph $G_{f_\theta}$ based on the contribution of the input example $x$. Algorithm~\ref{alg:normalization} only applies when ${n_{l_1,j}(x) > 0}$ since otherwise the ReLU would be its zero phase and all outgoing edges would be removed. Algorithm~\ref{alg:normalization} normalizes the weights such that negative weights are reset to 0 and positive weights are normalized so that overall sum remains the same. Given Definition~\ref{def:neural_data_graph}, we are ready to define the neural Ricci curvature, which is essentially the ORC applied to the neural data graph.

\begin{definition}[Neural Ricci Curvature]
Consider a NN $f_\theta$, an input example $x$ and a corresponding neural data graph $G_{f_\theta,x}$, as defined in Definition~\ref{def:neural_data_graph}. The NRC of a NN edge $(n_{l,i},n_{l+1,j})$ is defined as the ORC of the corresponding edge in $G_{f_\theta,x}$.
\end{definition}
\section{Evaluation}
\label{sec:evaluation}
The evaluation aims to demonstrate the following: given a NN, $\varepsilon$-robust examples with higher $\varepsilon$ tend to result in neural data graphs with fewer negative-NRC edges. Note that this result is orthogonal to the overall NN robustness -- even non-robust NNs exhibit some $\varepsilon$-robust examples with large $\varepsilon$. This finding implies that training NNs with fewer negative-NRC edges would be an effective method for robust training.

We evaluate the NRC concept on the MNIST dataset~\cite{lecun98}. MNIST consists of ${28\times 28}$ grayscale images of handwritten digits; there are 60,000 training images and 10,000 test images. We use two fully-connected NN architectures, [15,20] and [15,25,20,15], where notation $[K_1, \dots, K_L]$ means the NN has $L$ layers, with $K_i$ neurons in layer $i$. Each architecture is trained in three ways: 1)~using cross-entropy loss; 2)~using cross-entropy with weight decay regularization; 3)~using adversarial training~\cite{madry17}. For further evaluation, we also include a convolutional NN (CNN), trained with cross-entropy, with two convolutional layers (first layer has six ${6\times 6}$ kernels, stride of two; second layer has 16 ${6 \times 6}$ kernels, stride of two) and two fully-connected layers, [120,84]. The models' robust accuracies are reported in Table~\ref{tab:all_acc}.
All experiments were run on a 95-core machine with an NVIDIA A40 unit; calculating curvatures per example took on average 1.2s, 1.4s and 3.2s for the two-layer NN, four-layer NN and CNN, respectively.

\begin{figure*}[!th]
  \centering 
       \subfloat[Cross-entropy, label 0.]{
           \includegraphics[width=0.33\linewidth]{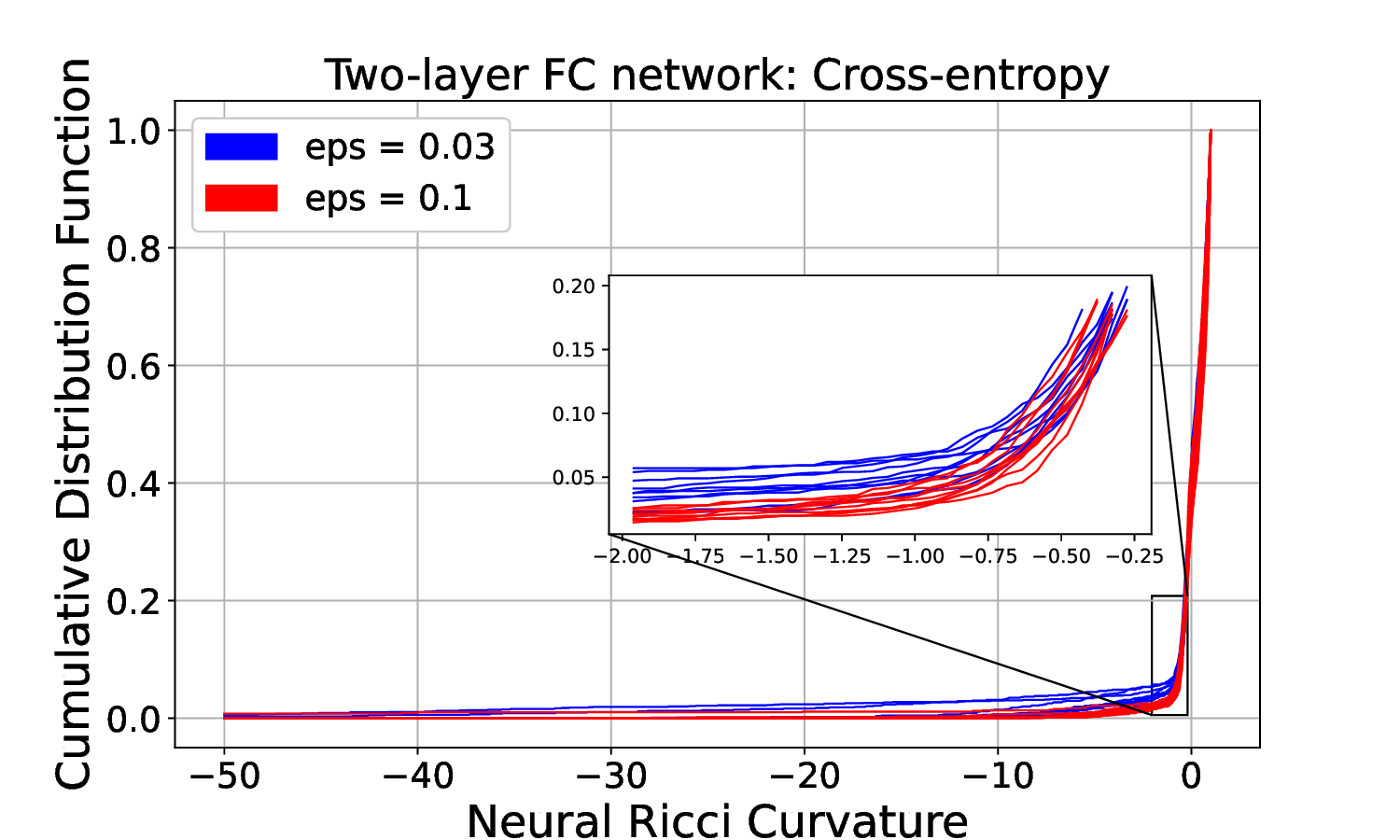}
           \label{fig:two_layer_st}
       }
       \subfloat[Cross-entropy + weight decay, label 2.]{
           \includegraphics[width=0.33\linewidth]{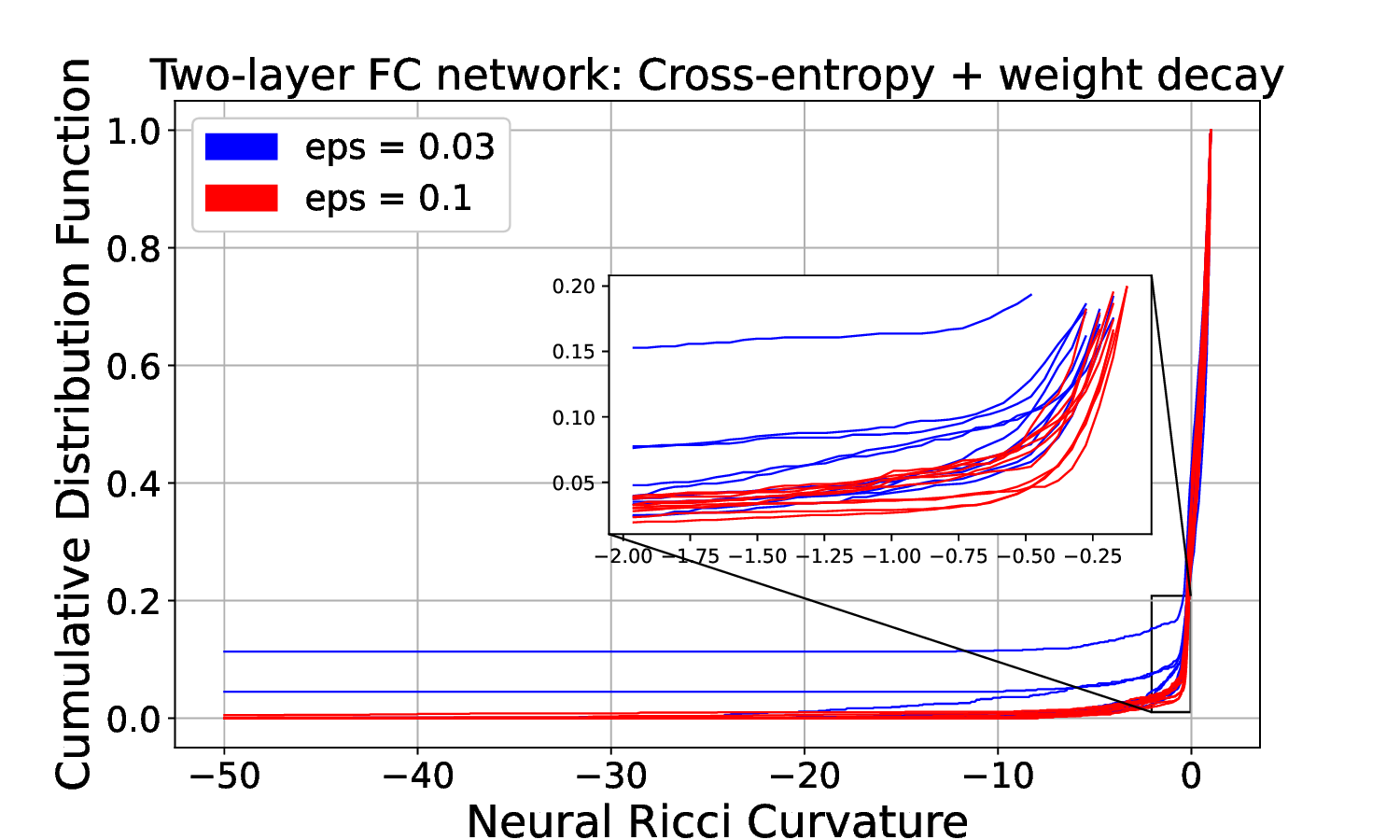}
           \label{fig:two_layer_wd}
       }
       \subfloat[Adversarial training, label 6.]{
           \includegraphics[width=0.33\linewidth]{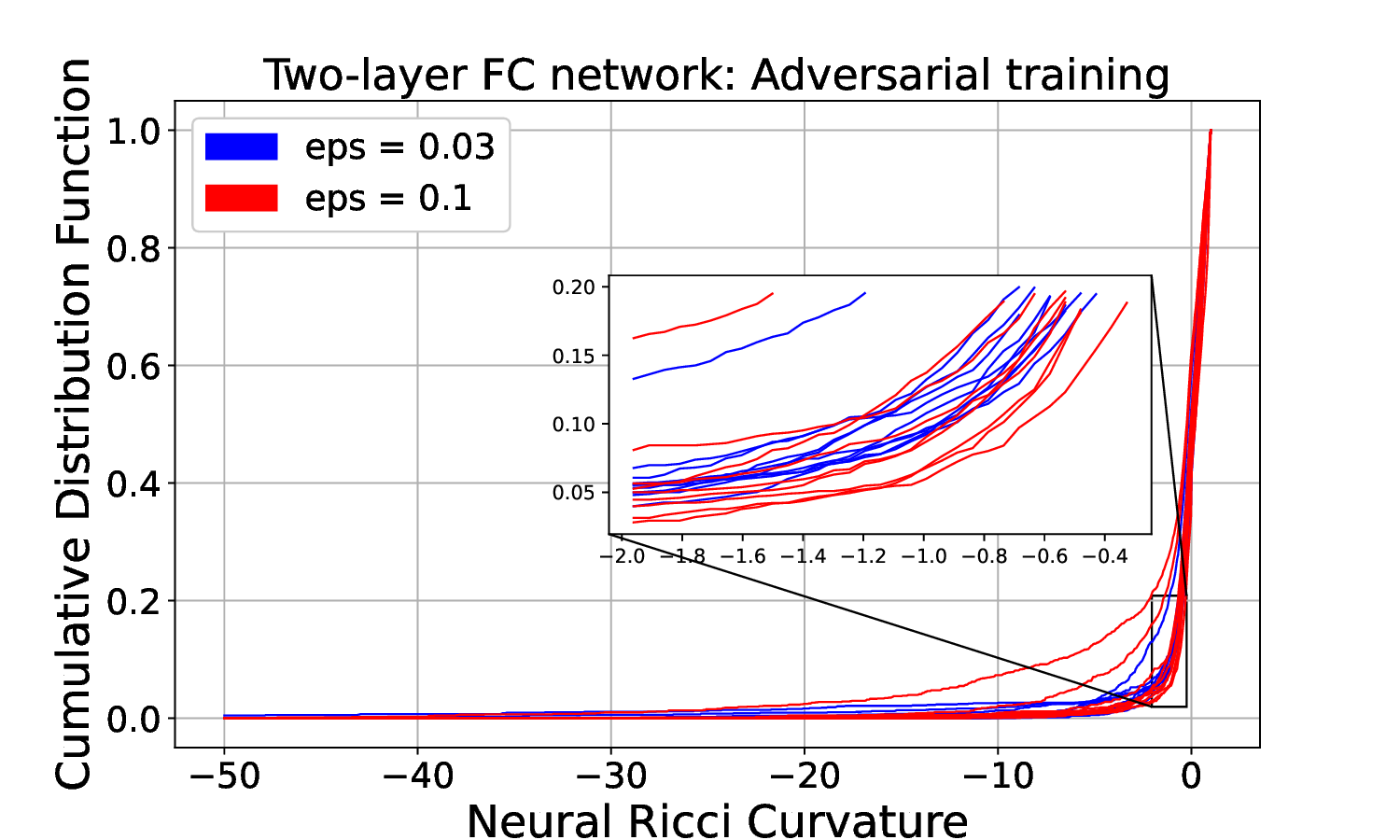}
           \label{fig:two_layer_at}
       }
       \vspace{-5px}
       \caption{CDF plots of 10 test examples each for two levels of $\varepsilon$, for the three two-layer NN setups. Note that the 0.03-robust examples are chosen such that they are 0.05-nonrobust.}
       \vspace{-22px}
  \label{fig:cdf}
\end{figure*}

To perform the evaluation, we use the test set to identify $\varepsilon$-robust and $\varepsilon$-nonrobust images for each setup, where $\varepsilon$ is incremented from $0.03$ to $0.2$. We calculate the NRCs for each resulting neural data graph and plot them using a cumulative distribution function (CDF). As shown in Fig.~\ref{fig:cdf}, the CDF grows faster for examples which are less $\varepsilon$-robust, i.e., those examples have more negatve-NRC edges, on average, than the graphs corresponding to more $\varepsilon$-robust examples. 

\begin{table}
\centering
\begin{tabular}{|l|c|c|c|c|}
\hline
NN setup  & $\varepsilon=0.03$ & $\varepsilon=0.07$ & $\varepsilon=0.1$ & $\varepsilon=0.2$ \\ \hline
[15,20], CE & 0.517& 0.044 & 0.005 & 0.000\\ \hline
[15,20], WD & 0.851 & 0.421&0.175 & 0.014  \\ \hline
[15,20], AT & 0.845&0.766 &0.685 & 0.270 \\ \hline
[15,25,20,15], CE & 0.471& 0.054 &0.007 &0.000 \\ \hline
[15,25,20,15], WD &0.801 & 0.311 & 0.109&0.001 \\ \hline
[15,25,20,15], AT & 0.862& 0.780 &0.692 & 0.253\\ \hline
CNN, CE & 0.939 & 0.725 &0.409 & 0.017 \\ \hline
\end{tabular}
\caption{Robust test accuracy (evaluated using the projected gradient descent attack~\cite{madry17}) for all setups: cross-entropy (CE), cross-entropy + weight decay (WD), adversarial training (AT).}
\label{tab:all_acc}
\vspace{-5px}
\end{table}

\begin{table}
\centering
\begin{tabular}{|l|c|c|c|c|}
\hline
NN setup  & $\varepsilon=0.03$ & $\varepsilon=0.07$ & $\varepsilon=0.1$ & $\varepsilon=0.2$ \\ \hline
[15,20], CE & 1.03 & 0.92 & 0.91& N/A\\ \hline
[15,20], WD & 1.01  & 1.07  & 0.90 & 0.81 \\ \hline
[15,20], AT & 1.28 & 1.27   & 1.26 & 1.30 \\ \hline
[15,25,20,15], CE & 1.23 & 1.20 & 1.08& N/A \\ \hline
[15,25,20,15], WD & 1.34 & 1.30 &1.24 & N/A\\ \hline
[15,25,20,15], AT & 1.55 & 1.57 & 1.55& 1.54\\ \hline
CNN, CE & 3.60 & 3.60 & 3.73& 3.66 \\ \hline
\end{tabular}
\caption{Average AUC over all labels for all considered setups: cross-entropy (CE), cross-entropy + weight decay (WD), adversarial training (AT).}
\label{tab:all_auc}
\vspace{-15px}
\end{table}

\begin{figure*}[!th]
  \centering 
       \subfloat[Cross-entropy.]{
           \includegraphics[width=0.33\linewidth]{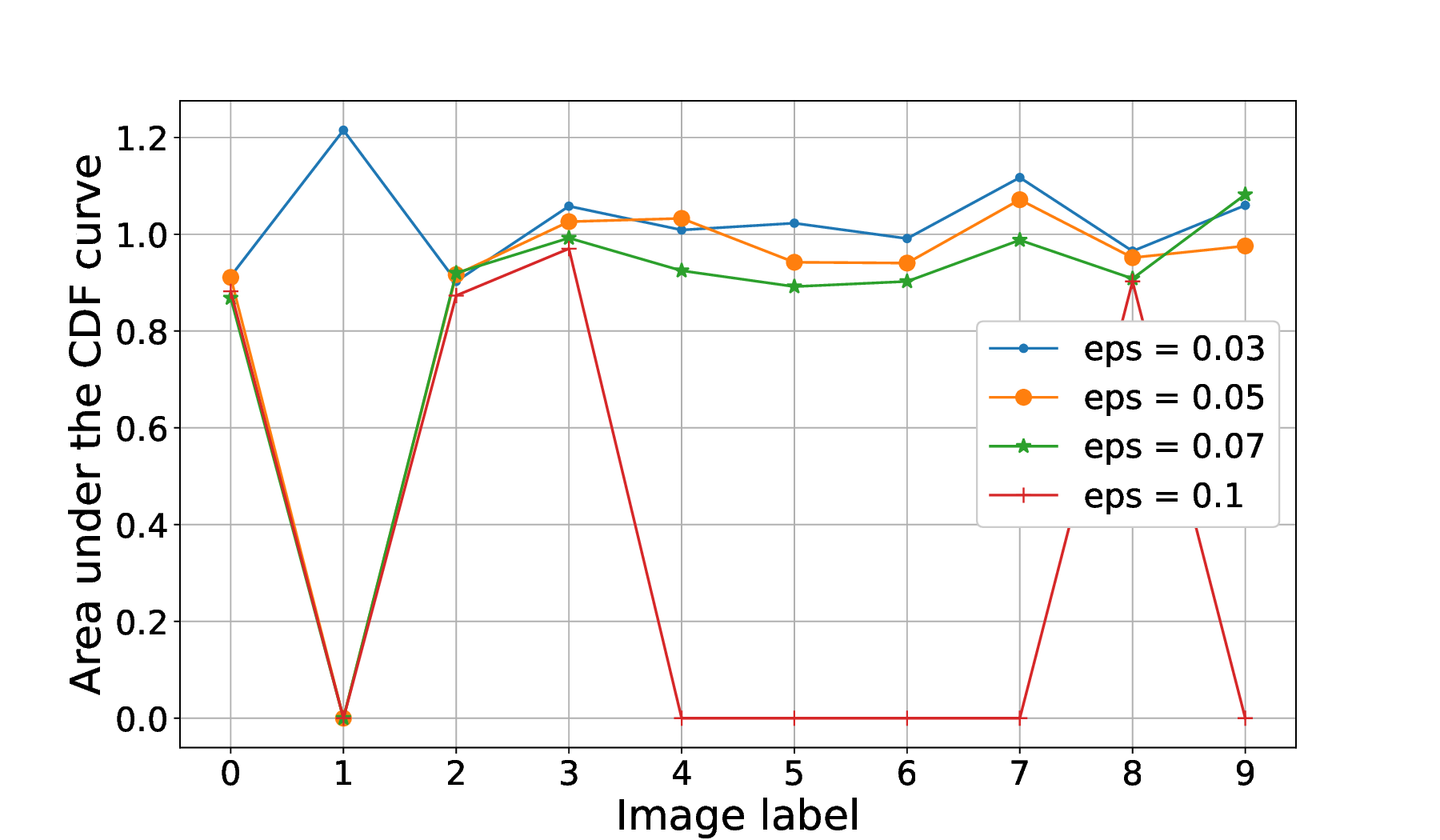}
           \label{fig:two_layer_st}
       }
       \subfloat[Cross-entropy + weight decay.]{
           \includegraphics[width=0.33\linewidth]{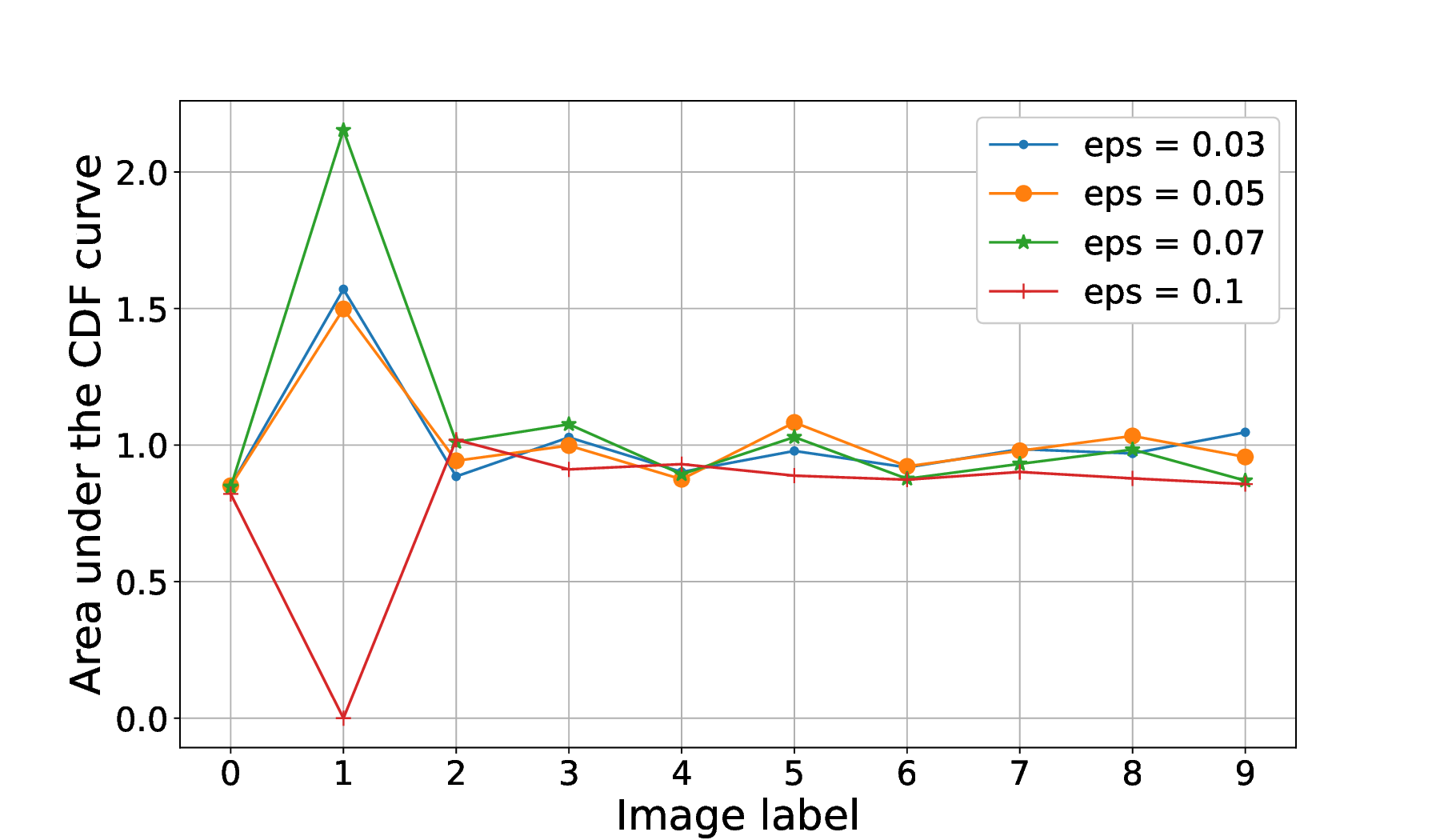}
           \label{fig:two_layer_wd}
       }
       \subfloat[Adversarial training.]{
           \includegraphics[width=0.33\linewidth]{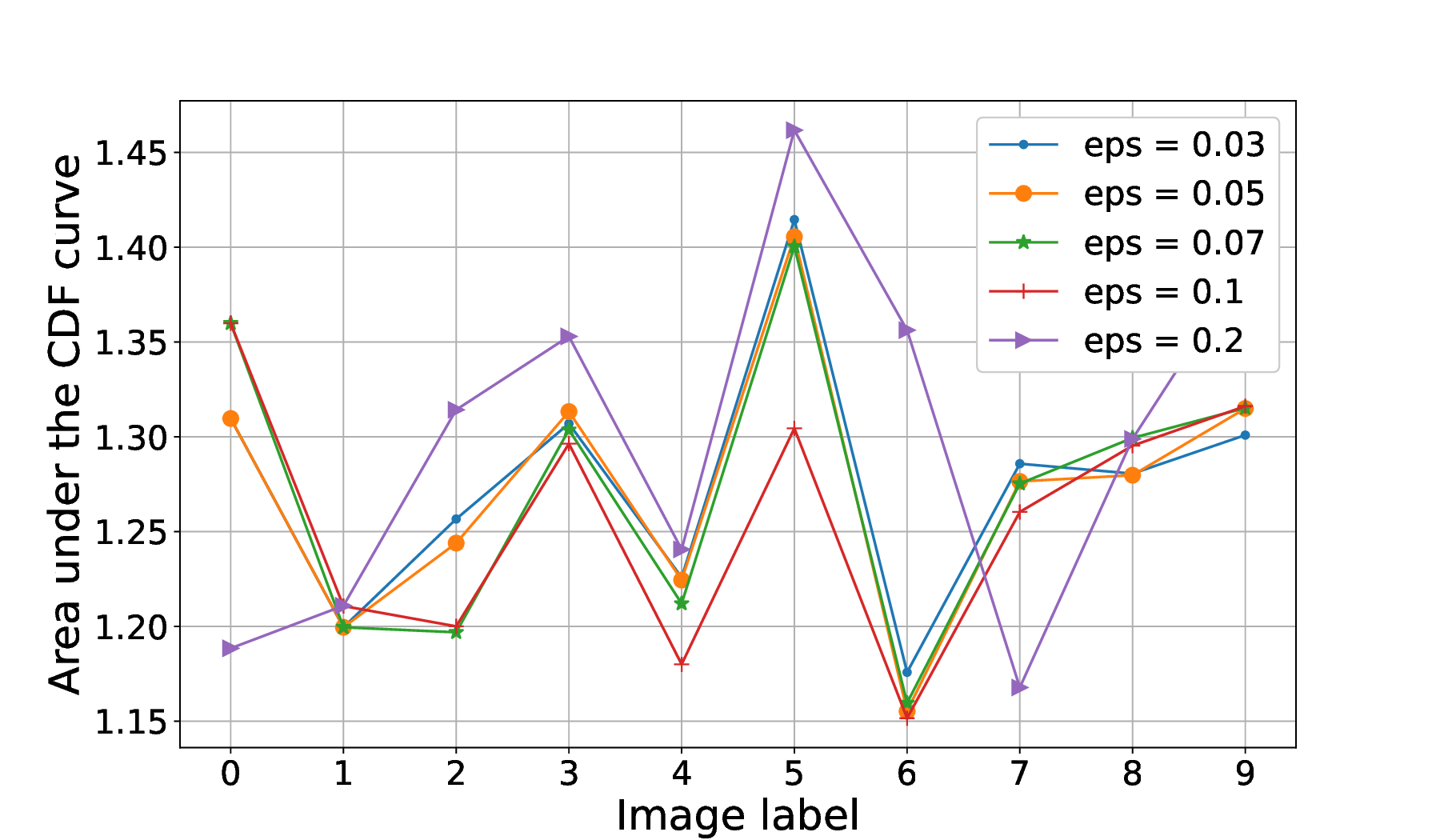}
           \label{fig:two_layer_at}
       }
       \vspace{-5px}
       \caption{AUC results, averaged over 70 test examples per label and per $\varepsilon$, for the three two-layer NN setups. Values of 0 mean that no robust examples could be found for that value of $\varepsilon$.}
       \vspace{-13px}
  \label{fig:auc}
\end{figure*}

For further evaluation, we also present the average area under the CDF curve (AUC) per label (averaged over 70 test examples). Fig.~\ref{fig:auc} provides the average results for the two-layer fully-connected (FC) NN. Fig.~\ref{fig:two_layer_st} very clearly demonstrates that the AUC decreases with larger $\varepsilon$, i.e., more $\varepsilon$-robust examples result in fewer negative-NRC edges on average. Although the benefit is less pronounced for the other setups, there is still a clear downwards trend as $\varepsilon$ is increased. Finally, Table~\ref{tab:all_auc} presents the AUC results for all setups. Once again, we emphasize that the same trend can be observed across all considered setups. Finally, we note that the CNN trends are less emphasized, which is likely due to the fact that convolutional layers have very few edges per node; we will explore this phenomenon at greater depth in future work.

\section{Conclusion}
\label{sec:conclusion}
This paper introduced neural data graphs and neural Ricci curvature, which provide an alternative way of analyzing NN robustness. We presented an evaluation on the MNIST dataset to demonstrate that more robust NNs (as well as more $\varepsilon$-robust examples) result in fewer negative-NRC edges. In future work, we will perform an evaluation over multiple datasets and robust training methods~\cite{brau2023robust,leino2021globally,zou2019lipschitz}, and will explore an approach for robust training, e.g., by regularizing training through a term that penalizes low-curvature edges.

\bibliography{references}
\bibliographystyle{IEEEtran}

\end{document}